\def\year{2019}\relax
\documentclass[letterpaper]{article} %DO NOT CHANGE THIS
\usepackage{aaai19}  %Required
\usepackage{times}  %Required
\usepackage{helvet}  %Required
\usepackage{courier}  %Required
\usepackage{url}  %Required
\usepackage{graphicx}  %Required
\usepackage{amsfonts}
\usepackage{amsmath}
\usepackage{amsthm}
\usepackage{algorithm}
\usepackage[noend]{algpseudocode}
\usepackage{color}
\usepackage[colorinlistoftodos]{todonotes}

\newcommand{\norm}[1]{\lVert#1\rVert}

\newcommand{\old}{\textrm{old}}
\newcommand{\E}{\mathbb{E}}
\newcommand{\bE}{\mathbb{E}}
\newcommand{\KL}{\mathrm{KL}}
\newcommand{\rsa}{r(s,a)}
\newcommand{\pias}{\pi(a|s)}
\newcommand{\piw}{{\pi_\theta}}
\newcommand{\piold}{{\pi_\old}}
\newcommand{\piwas}{{\pi_\theta(a|s)}}

\newcommand{\qpisa}{{Q^\pi(s,a)}}

\newcommand{\vpis}{{V^\pi(s)}}
\newcommand{\vphis}{{V_\phi(s)}}

\newcommand{\apisa}{{A^\pi(s,a)}}

\newcommand{\auxpi}{{\tilde\pi}}

\newcommand{\dw}{\nabla_\theta}
\newcommand{\deltapome}{\delta^\textrm{POME}}

\newtheorem{theorem}{Theorem}

\begin{document}
% The file aaai.sty is the style file for AAAI Press
% proceedings, working notes, and technical reports.
%
\title{Policy Optimization with Model-based Explorations}
\author{Feiyang Pan\thanks{These authors contributed equally in this work.}\textsuperscript{1,2}\thanks{The research was supported by the National Key Research and Development Program of China under Grant No. 2017YFB1002104, the National Natural Science Foundation of China under Grant No. 61602438, 91546122, 61573335, Guangdong provincial science and technology plan projects under Grant No. 2015 B010109005.}, Qingpeng Cai\footnotemark[1]\textsuperscript{3}\thanks{Pingzhong Tang and Qingpeng Cai were supported in part by the National Natural Science Foundation of China Grant 61561146398, a China Youth 1000-talent program and an Alibaba Innovative Research program.}, An-Xiang Zeng\textsuperscript{4}, Chun-Xiang Pan\textsuperscript{4}, Qing Da\textsuperscript{4}, \\ \bf \Large Hualin He\textsuperscript{4}, Qing He\textsuperscript{1,2}\footnotemark[2], Pingzhong Tang\textsuperscript{3}\footnotemark[3]\\
	\textsuperscript{1}Key Lab of Intelligent Information Processing of Chinese Academy of Sciences (CAS), \\Institute of Computing Technology, CAS, Beijing 100190, China. \{panfeiyang, heqing\}@ict.ac.cn\\
	\textsuperscript{2}University of Chinese Academy of Sciences, Beijing 100049, China. \\
	\textsuperscript{3}IIIS, Tsinghua University. cqp14@mails.tsinghua.edu.cn, kenshinping@gmail.com\\
	\textsuperscript{4}Alibaba Group. \{renzhong, xuanran\}@taobao.com, \{daqing.dq, hualin.hhl\}@alibaba-inc.com
}
\maketitle

\begin{abstract}
Model-free reinforcement learning methods such as the Proximal Policy Optimization algorithm (PPO) have successfully applied in complex decision-making problems such as Atari games. However, these methods suffer from high variances and high sample complexity. On the other hand, model-based reinforcement learning methods that learn the transition dynamics are more sample efficient, but they often suffer from the bias of the transition estimation. How to make use of both model-based and model-free learning is a central problem in reinforcement learning. 

In this paper, we present a new technique to address the trade-off between exploration and exploitation, which regards the difference between model-free and model-based estimations as a measure of exploration value. We apply this new technique to the PPO algorithm and arrive at a new policy optimization method, named Policy Optimization with Model-based Explorations (POME). POME uses two components to predict the actions' target values: a model-free one estimated by Monte-Carlo sampling and a model-based one which learns a transition model and predicts the value of the next state. POME adds the error of these two target estimations as the additional exploration value for each state-action pair, i.e, encourages the algorithm to explore the states with larger target errors which are hard to estimate. We compare POME with PPO on Atari 2600 games, and it shows that POME outperforms PPO on 33 games out of 49 games.
\end{abstract}

\section{Introduction}
Reinforcement Learning focuses on maximizing long-term return by interacting with the environment sequentially \cite{sutton1998reinforcement}. Generally, a reinforcement learning algorithm has two aspects, on the one hand, it estimates the state-action value function (also known as the Q-function), on the other hand, it optimizes or improves the policy to maximize its performance measure.

There are two classes of reinforcement learning methods, model-free and model-based methods. Model-free methods \cite{peters2006policy} estimate and iteratively update the state-action value with the rollout samples by Temporal Difference learning \cite{sutton1998reinforcement}. It is said that model-based methods maintain an approximate model including the reward functions and the state transitions, and then use the approximated rewards and transitions to estimate the value function. Model-based methods are more efficient than model-free methods \cite{li2004iterative,levine2013guided,montgomery2016guided,wahlstrom2015pixels,watter2015embed}  especially in discrete environments by reducing the sample complexity. Model-free methods are more generally applicable to continuous and complex control problems but may suffer from high sample complexity \cite{schulman2015trust,lillicrap2015continuous}.

Model-free methods directly use the immediate rewards and next states from rollout samples and estimate the long-term state-action value by Temporal Difference learning, for example, Sarsa or Q-learning \cite{sutton1998reinforcement}. Therefore, the target value is unbiased but may induce large variance due to the randomness of the transition dynamics or the off-policy stochastic exploration strategy. Model-based methods use the prediction of the immediate reward and the next state by its own belief of the environment. The belief of the agent, including the approximate reward function and transition model, is updated after receiving more signals from the environment. So the estimated target value in model-based methods is often biased due to the approximation error of the model, but it has low variance compared with model-free methods. Combining model-based and model-free learning has been an important question in the recent literature.

We aim to answer the following question: How to incorporate model-based and model-free methods for better control?

The Dyna-Q \cite{sutton1990integrated}, Normalized Advantage Function (NAF) algorithm \cite{gu2016continuous}, and Model-Ensemble Trust-Region Policy Optimization (ME-TRPO) algorithm \cite{kurutach2018model} use simulated experiences in a learned transition model to supplement the real transition samples. 
\cite{cai2018generalized} uses a convex combination of the two target values as the new target, following the insight that the ensemble could more accurate. But the large loss of model prediction would also lead to an incorrect update for policy iteration.

In this paper, we incorporate the two methods together to address the trade-off between exploration and exploitation (EE), by a simple heuristic that adds the discrepancy between both target values as a relative measure of the \textit{exploration value}, so as to encourage the agent to explore more difficult transition dynamics. 

The trade-off between EE is a fundamental and challenging problem in reinforcement learning because it is hard to evaluate the value of exploring unfamiliar states and actions. Model-free exploration methods are widely discussed over decades. To name some, Upper Confidence Bounds \cite{auer2002finite,li2010contextual,chen2017ucb} adds the upper confidence bound of the uncertainty of value estimation as an exploration value of actions. Thompson Sampling \cite{thompson1933likelihood} can also be understood as adding stochastic exploration value based on the posterior distribution of value estimation \cite{may2012optimistic}. However these model-free value-based methods often require a table to store the number of visits of each state-action pair so that the uncertainty of value estimation can be delivered \cite{tang2017exploration}, so they are not practical to handle problems with continuous state spaces such as Atari games or large-scale real-world problems such as mechanism design \cite{tang2017reinforcement} and e-commerce \cite{cai2018reinforcement,cai2018reinforcementaaai}. Model-based methods for deep RL can have better exploration based on maximization of information gain \cite{houthooft2016vime} or minimization of error \cite{pathak2017curiosity} about the agents' belief, however, may be unstable due to the difficulty of learning the transition dynamics.

In this paper, based on the discrepancy between model-free and model-based target values, we present a new algorithm named Policy Optimization with Model-based Exploration (POME), an exploratory modified version of the well-known algorithm Proximal Policy Optimization (PPO). Different from previous methods, we use the model-free sample-based estimator for the advantage function as the base for POME, which is already successfully used in various actor-critic methods including PPO. Then POME adds a centralized and clipped exploration bonus onto the advantage value, so as to encourage exploration as well as stabilize the performance.

Our intuition is that the discrepancy of target values of model-based and model-free methods can be understood as the uncertainty of the transition dynamics. A high exploration value means that the transition of the state-action pair is hard to estimate and needs to be explored. We directly add the difference to the advantage estimation as the exploration bonus of the state-action pair. So if the exploration value of a state-action pair is higher than average, POME will encourage the agent to visit it more frequently in the future so as to better learn the transition dynamics. If not, POME will reduce the chance of picking it and give the agent an opportunity to try other actions. 

We verify POME on the Atari 2600 game playing benchmarks. We compare our POME with the original model-free PPO algorithm and a model-based extended version of PPO. Experimental results show that POME outperforms the original PPO on 33 Atari games out of 49. We also tested two versions of POME, one with decaying exploration level and the other with non-decaying exploration level, which results in an interesting finding that the exploration bonus can improve the performance even in a long run.

\section{Background and Notations}
A Markov Decision Process (MDP) is defined as the tuple $(\mathcal{S}, \mathcal{A}, r, P, P_{0})$, where $\mathcal{S}$ is the (discrete or continuous) state space, $\mathcal{A}$ is the (discrete or continuous) action space, $r: \mathcal{S}\times\mathcal{A}\to \mathbb{R}$ is the (immediate) reward function, $P$ is the state transition model, and $P_{0}$ is the probability distribution for the initial state $s_{0}$. 

The goal of the agent is to find the policy $\pi^*$ from a restricted family of parametrized policy functions that maximizes its performance, 
\begin{equation}
\pi^* = \arg\max_{\pi\in\Pi} J(\pi),
\end{equation}
where $J(\pi)$ is the performance objective defined as
\begin{equation}
J(\pi) = \mathbb{E}_{s_0, a_0, s_1,\dots} \bigg[\sum_{t=0}^{\infty} \gamma^t r(s_t,a_t)\mid\pi, s_0\sim P_0\bigg],
\label{Def-of-J}\end{equation}
where $\gamma\in (0,1)$ is a discount factor that balances the short- and long-term returns. For convenience, let $\rho^\pi(s)$ denote the (unnormalized) discounted cumulated state distribution induced by policy $\pi$, 
\begin{equation}
\rho^\pi(s):=\sum_{t=0}^\infty \gamma^t \Pr(s_t=s|\pi, s_0\sim P_0)
\label{Def-of-rho}\end{equation}
then the performance objective can rewrite as 
$$J(\pi) = \mathbb{E}_{s\sim\rho^\pi, a\sim\pi}[r(s,a)].$$

Since the expected long-term return is unknown, the basic idea behind RL is to construct a tractable estimator to approximate the actual return. Then we can update the policy in the direction that the performance measure is guaranteed to improve at every step \cite{kakade2002approximately}. Let $Q^\pi(s,a)$ denote the expected action value of action $a$ at state $s$ by following policy $\pi$, i.e.
\begin{equation}
\qpisa=\mathbb{E}\bigg[\sum_{t=0}^{\infty}\gamma^{t} r(s_t,a_t)| s_0=s, a_0=a,\pi\bigg].
\label{Def-of-Q}\end{equation}
And we write the state value $\vpis$ and the advantage action value $\apisa$ as 
\begin{align}
&\vpis=\int_{\mathcal{A}}\pias \qpisa \mathrm{d}a,\\ 
&\apisa=\qpisa-\vpis.
\end{align}
So the state value is defined as the weighted average of action values, and the advantage function provides a relative measure of value of each action. 

Since we have the Bellman equation
\begin{equation}
\qpisa = \rsa+\gamma\E_{s'}[V^\pi(s')],
\end{equation}
the advantage function can also be written as 
\begin{equation}
\apisa=\rsa+\gamma\E_{s'}[V^\pi(s')]-\vpis.
\end{equation}

In practice, estimations of these quantities are used to evaluate the policy and to guide the direction of policy updates. 

\subsection{Policy Optimization Methods}

Policy Gradient (PG) methods \cite{sutton2000policy} compute the gradient of $J(\pi_\theta)$ w.r.t policy parameters $\theta$ and then use gradient ascent to update the policy. The well-known form of policy gradient writes as follow
\begin{equation}
\dw J(\piw) = \bE_{s\sim \rho^\piw,a\sim\piw} \big[A^{\piw}(s,a) \dw \log\piwas \big].
\label{pg}\end{equation} 

Instead of directly using the policy gradients, trust region based methods use a surrogate objective to perform policy optimization. For the reason that the trajectory distribution is known only for the prior policy $\piold$ before the update, trust region based methods introduce the following local approximation to $J(\piw)$ for any untested policy $\piw$, as
\begin{equation}
L_{\piold}(\piw)=J(\piold) + \bE_{s\sim \rho^{\piold},a\sim\piw}[\piw(a|s) A^{\piold}(s,a)].
\end{equation}
It is proved \cite{kakade2002approximately,schulman2015trust} that if the new policy $\piw$ is close to the $\piold$ in the sense of the KL divergence, there is a lower bound of the long-term rewards of the new policy. For the general case, we have the following result:
\begin{theorem}[Schulman et al. 2015]
Let $D^{\max}_\KL(\piold,\piw)=\max_s\KL[\piold(\cdot|s) \Vert \piw(\cdot|s)]$ and $A_{\max}= \max_s |\bE_{a\sim \piw}[A^\piold(s,a)]|$. The performance of the policy $\auxpi$ can be bounded by
\begin{equation}
J(\piw)\geq L_\piold(\piw) - \frac{2\gamma A_{\max}}{(1-\gamma)^2}D^{\max}_\KL(\piold,\piw).
\end{equation}
\label{trpo-theorem}\end{theorem}
By following Theorem \ref{trpo-theorem}, there is a category of policy iteration algorithms with monotonic improvement guarantee, known as conservative policy iteration \cite{kakade2002approximately}. Among them, Trust Region Policy Optimization (TRPO) is one of the most widely used baselines to optimize parametric policies to solve complex or continuous problems. The optimization problem w.r.t the new parameter $\theta$ is
\begin{align}
\max_\theta &\quad\bE_{s\sim \rho^{\piold},a\sim\piold}\bigg[\frac{\piwas}{\piold(a|s)} A^{\piold}(s,a) \bigg]\label{trpo-obj-original}\\
\mbox{s.t.} &\quad \KL[\piold(\cdot|s) \Vert \pi_\theta(\cdot|s)]\leq \delta_{\KL}\,\,\mbox{for all}\,s.\notag
\end{align} 
where $\delta_{\KL}$ is a hard constraint for the KL-divergence. In practice, the hard constraint of KL-divergence is softened by an expectation over visited states and moved to the objective with a multiplier $\beta$ so that it becomes an unconstrained optimization problem, i.e.
\begin{equation}
\max_\theta\bE\bigg[\frac{\piwas}{\piold(a|s)} A^{\piold}(s,a) 
-\beta\KL[\piold(\cdot|s) \Vert \pi_\theta(\cdot|s)]\bigg],
\label{trpo-obj}\end{equation}
which stabilise standard RL objectives. 

Proximal Policy Optimization (PPO) \cite{schulman2017proximal} can be view as a modified version of TRPO, which mainly uses a clipped probability ratio in the objective to avoid excessively large policy updates. It mainly replaces the term inside the expectation in (\ref{trpo-obj-original}) with 
\begin{equation}\begin{split}
l^\textrm{PPO}(\theta;s,&a)
=\min\big\{\frac{\piwas}{\piold(a|s)} A^{\piold}(s,a),\\
&\mathrm{clip}\big(\frac{\piwas}{\piold(a|s)}, 1-\epsilon, 1+\epsilon\big) A^{\piold}(s,a)\big\}.
\end{split}\end{equation}
PPO is said to be empirically more sample efficient than TRPO, and is made one of the most frequently used baselines in various kinds of deep RL tasks.
\subsection{Policy Evaluation with Function Approximation}
In the previous part, we show several objectives for policy optimization. The objectives commonly need to compute an expectation over $(s,a)$ pairs obtained by following the current policy and an estimator of advantage values. 

In on-policy methods such as TRPO and PPO, the expectation is approximated with the empirical average over samples induced by the current policy $\piold$. Therefore the policy updates in a similar manner to the stochastic gradient descent as in modern supervised learning.
\subsubsection{Model-free value estimation}
Model-free methods estimate the value functions by the samples in the rollout trajectory $(s_0, a_0, r_0, s_1, \dots, )$. By following the Bellman equation, we can estimate the action or advantage value by Temporal Difference learning. When referring to the quantities at time step $t$ of a trajectory, we will use $ Q_t$, $V_t$, and $A_t$ for short to denote $Q(s_t,a_t)$, $V(a_t)$, and $A(s_t,a_t)$ respectively.

Policy-based methods, as described in the previous part, typically need to use the advantage value $A_t$. It is discussed in \cite{schulman2015trust,wang2016sample} that it can be one of the following estimators without introducing bias: the state-action value $Q_t$, the discounted return of the trajectory started from $(s_t,a_t)$, the one-step temporal difference error (TD error) 
\begin{equation}
\delta_t = r_t+\gamma V_{t+1}-V_t,
\label{td-err}\end{equation}or the k-step cumulated temporal difference error as used in the actual algorithm of PPO
\begin{equation}
\delta_t^{k}:=\sum_{j\geq 0}^{k-1} (\gamma\lambda)^j\,\delta_{t+j},
\label{td-err-k}\end{equation} where $\lambda$ is a discount coefficient to balance future errors. When $\lambda=1$, the estimator naturally becomes the advantage approximation in the asynchronous advantage actor critic algorithm (A3C) \cite{mnih2016asynchronous} $A_{\textrm{A3C}}(s_t, a_t)=\sum_{j\geq 0}^{k-1} \gamma^j r_{t+j}+\gamma^k V_{t+k}- V_t$. It is known that these estimators would not introduce bias to the policy gradient, but they have different variances. 

Moreover, in deep RL, we often approximate the value functions with neural networks, which introduces additional biases and variances \cite{mnih2015human,schulman2015trust,wang2016sample,schulman2017proximal}. For example, to compute the temporal difference errors, the state value $\vpis$ is always approximated with a neural network $\vphis$, where $\phi$ is the network parameter. Then if the one-step TD error is computed with on-policy samples, it becomes a loss function w.r.t the parameters $\phi$,
\begin{equation}
\delta_t(\phi)=Q^*_{f,t}-V_\phi(s_t),
\label{td-err-phi}\end{equation}
where $Q^*_{f,t}$ denotes the model-\textit{f}ree target value\begin{equation}
Q^*_{f,t}=r_t+\gamma V_\phi(s_{t+1}).
\label{target-model-free}
\end{equation}
Therefore in many algorithms the value parameters are updated by minimizing the error on samples using stochastic gradient descent.

The methods discussed so far can be categorized into model-free methods, because the immediate rewards and the state transitions are sampled from the trajectories. 

\subsubsection{Model-based value estimation}For the model-based case, additional function approximation should be used to estimate the reward function and the transition function, i.e. 
\begin{equation}
\hat r(s_t, a_t)\approx r_t\quad\mbox{ and }\quad\hat T(s_t,a_t)\approx s_{t+1}. 
\end{equation}
In continuous problems the form of $\hat r(s_t, a_t)$ and $\hat T(s_t, a_t)$ can be  neural networks, which are trained by minimizing some error measures between the model predictions and the real $r_t$ and $s_{t+1}$. Here the symbols for parameters are omitted for simplicity. When using the mean squared error, we can define the loss function to these two networks
\begin{align*}
& L_r = \bE \big[r_t - \hat r(s_t, a_t) \big]^2 \mbox{, }L_T = \bE\norm{s_{t+1} - \hat T(s_t, a_t)}_2^2
\end{align*}
which can be minimized using stochastic gradient descent w.r.t the network parameters.
To make use of the model-based estimators, similar to the model-free case, we have the model-based TD error 
\begin{equation}
\tilde \delta_t(\phi)=Q^*_{b,t} - V_\phi(s_t)
\label{td-err-phi-mb}\end{equation}
as an alternative to $\delta_t$ in (\ref{td-err-phi}),
where $Q^*_{b,t}$ denotes the model-\textit{b}ased target value
\begin{equation}
Q^*_{b,t}=\hat r(s_t, a_t) + \gamma V_\phi(\hat T(s_t,a_t)).
\label{target-model-based}\end{equation}

To solve complex tasks with high-dimensional inputs, model-free deep RL methods can often learn faster comparing to model-based methods especially in the beginning, mainly because it has fewer parameters to learn. 
%However it lacks interpretability because it can only provides value estimation for states and actions but can not tell why since it has no idea about the reward function nor the state transitions.

\section{Policy Optimization with Model-based Explorations}
In this section, we propose an extension of the trust region based policy optimization method with the exploration heuristics by making use of the difference between the model-free and model-based value estimations. 

Let us think of the TD error as a subtract between the target value and the current function approximation. The target value for model-free and model-based learning is written in (\ref{target-model-free}) and (\ref{target-model-based}). Here we define the \textit{discrepancy of targets} of state-action pair $(s_t, a_t)$ as 
\begin{align}
&\epsilon_t=|Q^*_{b,t}-Q^*_{f,t}|\\
&=|r_t+\gamma V(s_{t+1}) - \hat r(s_t, a_t) -\gamma V(\hat T(s_t,a_t))|.
\end{align}
For our method, we use this error as an additional exploration value for the pair $(s_t, a_t)$. The proposed methods is named Policy Optimization with Model-based Exploration (POME). So the resulted TD error used in POME is
\begin{equation}
\label{deltapome}
\deltapome_t=\delta_t + \alpha(\epsilon_t - \bar\epsilon),
\end{equation}
where $\alpha>0$ is a coefficient decaying to $0$ and $\bar\epsilon$ is used to shift the exploration bonus to zero mean over samples from the same batch. 

\subsection{Insights}
The basic idea behind POME is to add an exploration bonus to the state-action pairs where there is a discrepancy between the model-based and model-free Q-values. 

The intuition is that the discrepancy between both Q-values would be small if the agent is ``familiar'' with the transition dynamics, for example, when the probability distribution of the next state concentrates on one deterministic state and at the meanwhile the next state has already been visited several times. On another hand, the error would be large if the agent is uncertain about what is going on, for example, if it has trouble predicting the next state or even it has never visited the next state yet. Therefore we think of the \textit{discrepancy of targets} as a measure of the uncertainty of the long-term value estimation. 

By the update rule of policy iteration methods (\ref{pg}) and (\ref{trpo-obj-original}), the chance that the policy selects the action with larger advantage estimation would be higher after a few updates. When starting to learn, the discrepancy of both target values can be high, which means that the transition dynamic is hard to estimate, thus we need to encourage the agent to explore. After training properly for a while, we tend to schedule $\alpha$ to a tiny value approaching $0$, which means that the algorithm can asymptotically reach the model-free objective.

By the definition of (\ref{deltapome}), we let the discrepancy of targets serve as an additional exploration value of actions, along with the coefficient $\alpha$ to address the trade-off between exploration and exploitation. Note that, the exploration bonus on some state-action pairs will then be propagated to others so that our method guides exploration to these ``interesting'' parts of the state-action space.
\subsection{Techniques}
With the definition of the discrepancy of targets $\epsilon_t$ and the new exploratory target value $\deltapome_t$, we still need some techniques to build an actual policy optimization algorithm that is stable and efficient.

We first discuss why we use $\bar\epsilon$ for zero-mean normalization. Consider an on-policy algorithm that optimizes its policy using the samples $\{(s_t, a_t, r_t, s_{t+1})\}$ visited by following the old policy, i.e. $s\sim \rho^{\piold}, a\sim\piold$. Conventional methods estimate the advantage values on these state-action pairs to guide the next step of policy iteration. It means that, for a state $s$, if there exists a certain action $a$ that has never been taken before, it has to wait for a moment that the algorithm gives negative advantage values to other actions at the same state, otherwise it will never get a chance of being chosen. However, when we use POME to give biased advantage estimations by adding another term $\alpha (\epsilon_t - \bar\epsilon)$ where $\bar\epsilon$ is used to normalize $\epsilon_t$ to zero mean, it naturally reduces the chance of exploiting familiar actions and encourages to pick unfamiliar actions which either have uncertain results or have little chance to be picked before. For the actual algorithm, $\bar\epsilon$ is calculated as the median of $\epsilon_t$ of on-policy samples because the median of samples is more robust to outliers than the mathematical average.

Next, we show a practical trick to stabilize the exploration. Since the discrepancy of targets $\epsilon_t$ is an unbounded positive value, it can be extremely large when the agent arrives at a totally unexpected state for the first time, so the algorithm would be unstable. So we clip the exploration value to a certain range to stabilize the algorithm. By following Theorem {\ref{trpo-theorem}}, it is necessary to reduce the error when estimating $\hat A(\cdot)$ to guarantee monotonic improvements of policy iteration. So we clip the discrepancy of targets to the same range of the model-free TD error, by replacing (\ref{deltapome}) with
\begin{equation}
\deltapome_t=\delta_t + \alpha\,\textrm{clip}(\epsilon_t - \bar\epsilon, -|\delta_t|, |\delta_t|).
\label{deltapome-clip}\end{equation}
\begin{algorithm}
\caption{Policy Optimization with Model-based Explorations (single worker)}\label{alg:POME}
\begin{algorithmic}[1]
\State Input $k$\,\,\,\,\,// number of time steps per trajectory
\State Initialize $w,\phi,\theta_{\hat r},\theta_{\hat T}$\quad // parameters of $\pi,V,\hat r,\hat T$ 
\State Initialize $\alpha$\quad // coefficient for exploration
\While{not end}
    \State Sample a trajectories from the environment
    \For{$t=1$ to $k$}
    	\State $Q^{*}_{f,t} = r_t+\gamma V_\phi(s_{t+1})$
		\State $Q^{*}_{b,t} = \hat r(s_t, a_t)+\gamma V_\phi(\hat T(s_t, a_t))$
        \State $\epsilon_{t} = |Q^{*}_{b,t} - Q^{*}_{f,t}|$ 
    \EndFor
    \State $\bar\epsilon = \textrm{Median}(\epsilon_{1},\dots,\epsilon_{m=k})$
    \For {$t=k$ to $1$}
        \State $Q^{*}_{\textrm{POME},t}= Q^{*}_{f,t}+ \alpha\,\textrm{clip}(\epsilon_{t}-\bar\epsilon, -|\delta_t|, |\delta_t|)$
        \State $\deltapome_t(\phi)=Q^{*}_{\textrm{POME},t}-V_{\phi}(s_{t})$
        \If{ $t==m$ }
        {$\hat A^{\textrm{POME}}_{t}=\deltapome_t(\phi)$}
        \Else { $\hat A^{\textrm{POME}}_{t}=\deltapome_t(\phi)+\hat A^{\textrm{POME}}_{t+1}$} \EndIf
    \EndFor
    \State Update $\theta$ to maximize the surrogate objective (\ref{pome-obj})
    \State Update $\phi$ to minimize $L_v(\phi)$  
    \State Update $\theta_{\hat{r}}, \theta_{\hat{T}}$ to minimize $L_r(\theta_{\hat r})$ and $L_T(\theta_{\hat T})$
    \State Decay the coefficient $\alpha$ towards $0$ 
\EndWhile
\end{algorithmic}
\end{algorithm}

Formally, the optimization problem for POME is 
\begin{equation}
\max_\theta\bE\bigg[l^\textrm{POME}(\theta;s,a)
-\beta\KL[\piold(\cdot|s) \Vert \pi_\theta(\cdot|s)]\bigg],
\label{pome-obj}\end{equation}
where
\begin{equation}\begin{split}
l^\textrm{POME}&(\theta;s,a)
=\min\big\{\frac{\piwas}{\piold(a|s)} \hat A^{\textrm{POME}}(s,a),\\
&\mathrm{clip}\big(\frac{\piwas}{\piold(a|s)}, 1-\epsilon, 1+\epsilon\big)\hat A^{\textrm{POME}}(s,a)\big\}.
\end{split}\end{equation}
Following the one-step temporal difference error defined in (\ref{deltapome-clip}), we can use the k-step cumulated error to estimate the advantage value of state-action pairs along the trajectory,  
\begin{equation}
\hat A^{\textrm{POME}}_{t}:=\delta_t^{k,\textrm{POME}}=\sum_{j\geq 0}^{k-1} (\gamma\lambda)^j\,\delta_{t+j}^{\textrm{POME}}.
\label{adv-pome}\end{equation}
Therefore it is easy to compute by simply replacing the advantage estimation. The overall sample-based objective for policy parameters $\theta$ is
\begin{equation}\begin{split}
L^\textrm{POME}(\theta)=\frac{1}{T}\sum_{t=1}^T&\bigg[l^\textrm{POME}(\theta;s_t,a_t)\\
&-\beta\KL[\piold(\cdot|s_t) \Vert \pi_\theta(\cdot|s_t)]\bigg].
\end{split}\end{equation}
The loss function for value estimator therefore becomes 
\begin{equation}
L_v(\phi):=\frac{1}{T}\sum_{t=1}^T[\hat A^{\textrm{POME}}_t+ V_t - V_\phi(s_t)]^2.
\end{equation}

Now we present our new algorithm named Policy Optimization with Model-base Exploration (POME) by incorporating the techniques above, as shown in Algorithm 1. 

\section{Experiments}
\subsection{Experimental Setup}
We use the Arcade Learning Environment \cite{bellemare2013arcade} benchmarks along with a standard open-sourced PPO implementation \cite{baselines}.

We will evaluate two versions of POME in this section. For the first version, to guarantee that POME asymptotically approximates the original objective, we linearly anneal the coefficient $\alpha$ from $0.1$ to $0$ over the training period. {For the second version, we fix the value of $\alpha$ as 0.1}. The first version is used to show that the heuristic of POME helps fast learning because it mainly influences the algorithm in the beginning phase. The second version additionally shows that, even though the exploration value added to the estimation of advantage value introduces additional bias, the bias would not damage the performance too much. 

We test on all 49 Atari games. Each game is run for 10 million timesteps, over 3 random seeds. The average score of the last 100 episodes will be taken as the measure of performance. The experimental results of PPO in the comparison of Table 1 and Table 2 are borrowed from the original paper \cite{schulman2017proximal}.

\subsection{Implementation Details}
For all the algorithms, the discount factor $\gamma$ is set to $0.99$ and the advantage values are estimated by the k-step error $(\ref{td-err-k})$ where the horizon $k$ is set to be $128$. We use 8 actors (workers) to simultaneously run the algorithm and the mini-batch size is $128\times 8$.
 
For the basic network structures of PPO and POME, we use the same actor-critic architecture as \cite{mnih2016asynchronous,schulman2017proximal}, with shared convolutional layers and separate MLP layers for the policy and the value network. 

Since POME is an extended version of PPO by incorporating model-based value estimations, the implementation of POME involves two additional parts: fitting the model and adding the discrepancy of targets. Note that PPO and POME were always given the same amount of data during training.

\begin{figure}[!htb]
    \centering
    \includegraphics[width=\linewidth]{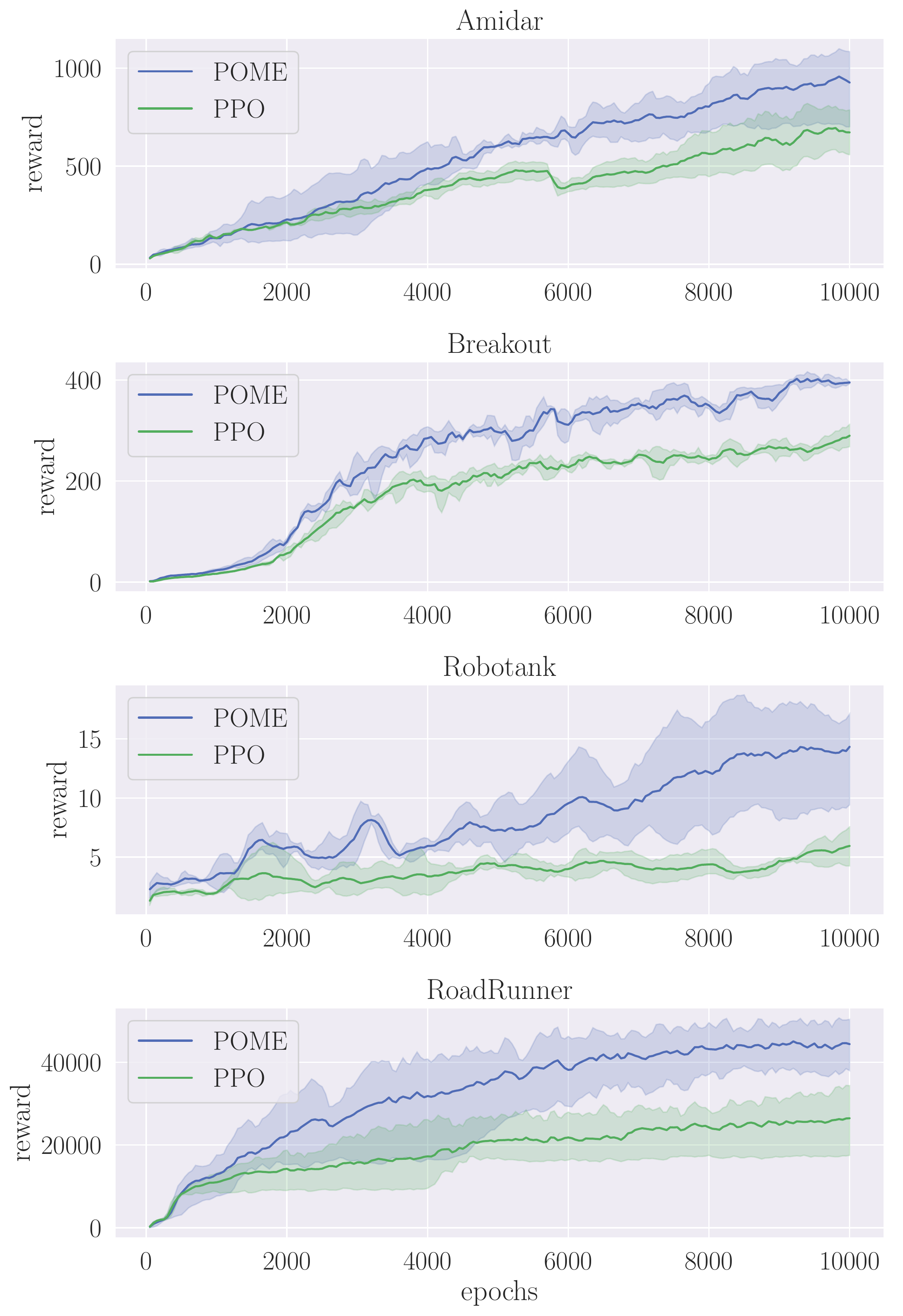}
    \caption{\small Comparison of POME against PPO on Atari games, training for 10M timesteps, over 3 random seeds.}\label{fig:ppo}
\end{figure}

To estimate the model, due to the fact that learning the state dynamics is more important than learning the rewards, we use a convolutional neural network with one hidden layer to fit the state transition. The inputs of the transition network have two parts, the state (images of four frames) and the action (a discrete number). Before being fed into the model, the images are scaled to the range $[0,1]$, and the action is one-hot encoded. We concatenate the one-hot encoding of the action to the states' images to form the inputs of the transition model. We use the \textit{sigmoid} activation for the outputs of the transition network. After that, we can finally compute the loss function of the transition model between the scaled images of the next state and the outputs.

\begin{table}[!t]
\small
\caption{Comparison between the original PPO and POME with decaying exploration coefficient. The scores of PPO are from the original paper \cite{schulman2017proximal}}
\begin{center}
\begin{tabular}{l|r|r}
{\sc Games} & {\sc PPO} & {\sc POME}\\
\hline
Alien              &         1850.3 &    {\bf1897.0} \\
Amidar             &          674.6 &     {\bf943.9} \\
Assault            &         4971.9 &    {\bf5638.6} \\
Asterix            &         4532.5 &    {\bf4989.2} \\
Asteroids          &    {\bf2097.5} &         1737.6 \\
Atlantis           & {\bf2311815.0} &      1941792.3 \\
BankHeist          &    {\bf1280.6} &         1241.7 \\
BattleZone         &   {\bf17366.7} &        15156.7 \\
BeamRider          &         1590.0 &    {\bf1815.7} \\
Bowling            &           40.1 &      {\bf58.3} \\
Boxing             &      {\bf94.6} &           92.9 \\
Breakout           &          274.8 &     {\bf411.8} \\
Centipede          &    {\bf4386.4} &         2921.6 \\
ChopperCommand     &         3516.3 &    {\bf4689.0} \\
CrazyClimber       &       110202.0 &  {\bf115282.0} \\
DemonAttack        &        11378.4 &   {\bf14847.1} \\
DoubleDunk         &          -14.9 &      {\bf-6.8} \\
Enduro             &          758.3 &     {\bf835.3} \\
FishingDerby       &           17.8 &      {\bf21.1} \\
Freeway            &           32.5 &      {\bf33.0} \\
Frostbite          &     {\bf314.2} &          272.9 \\
Gopher             &         2932.9 &    {\bf4801.8} \\
Gravitar           &          737.2 &     {\bf914.5} \\
IceHockey          &      {\bf-4.2} &           -4.5 \\
Jamesbond          &     {\bf560.7} &          507.2 \\
Kangaroo           &    {\bf9928.7} &         2511.0 \\
Krull              &         7942.3 &    {\bf8001.1} \\
KungFuMaster       &        23310.3 &   {\bf24570.3} \\
MontezumaRevenge   &      {\bf42.0} &            0.0 \\
MsPacman           &    {\bf2096.5} &         1966.5 \\
NameThisGame       &    {\bf6254.9} &         5902.2 \\
Pitfall            &          -32.9 &      {\bf-0.3} \\
Pong               &           20.7 &      {\bf20.8} \\
PrivateEye         &           69.5 &     {\bf100.0} \\
Qbert              &        14293.3 &   {\bf15712.8} \\
Riverraid          &         8393.6 &    {\bf8407.9} \\
RoadRunner         &        25076.0 &   {\bf44520.0} \\
Robotank           &            5.5 &      {\bf14.6} \\
Seaquest           &         1204.5 &    {\bf1789.7} \\
SpaceInvaders      &          942.5 &     {\bf964.2} \\
StarGunner         &        32689.0 &   {\bf44696.7} \\
Tennis             &     {\bf-14.8} &          -15.5 \\
TimePilot          &    {\bf4232.0} &         4052.0 \\
Tutankham          &     {\bf254.4} &          199.8 \\
UpNDown            &        95445.0 &  {\bf181250.4} \\
Venture            &              0 &       {\bf2.0} \\
VideoPinball       &   {\bf37389.0} &        33388.0 \\
WizardOfWor        &         4185.3 &    {\bf4301.7} \\
Zaxxon             &         5008.7 &    {\bf6358.0} \\
\end{tabular}
\end{center} 

\label{table1}\end{table}

In the actual implementation, POME use a unified objective function in order to simplify the computation
\begin{equation}
L=-L^\textrm{POME}(\theta)+c_vL_v(\phi)+c_TL_T(\theta_{\hat T}),
\end{equation} 
which is optimized by the Adam gradient descent optimizer \cite{kingma2014adam} with learning rate $2.5\times 10^{-4}\times f$, where $f$ is a fraction linearly annealed from 1 to 0 over the course of learning, and $c_v$, $c_T$ are coefficients for tuning the learning rate of the value function and the transition function. In our experiments we set these coefficients to $c_v=1$ and $c_T=2$.

\subsection{Comparison with PPO}
Table 1 compares POME with decaying coefficient $\alpha$ against the original PPO, on the averaged scores of the last 100 episodes of algorithms with each environment. In Table 1, we see that, among the 49 games, POME with decaying coefficient $\alpha$ outperforms PPO in 32 games at the last 100 episodes. 

The learning curves of four representative Atari games is shown in Figure 1. It shows that, in these environments, POME outperforms PPO over the entire training period, which indicates that it achieves fast learning and validate the power of our exploration technique by using the discrepancy of targets as exploration value.
\subsection{Additional experimental results}
We now investigate two questions: 
(1) how would POME perform if we do not tune the coefficient to $0$? 
(2) how would the direct model-based extension of PPO perform?

For the first question, we set up the experiment to see if the exploration value used in POME would damage the performance in a long run. The coefficient $\alpha$ is now set to $0.1$ for the entire training period. 
Secondly, we implement a model-based extension of PPO with the same architecture of the transition network as POME and replacing the target value with the model-based target value (\ref{target-model-based}), so the agent can perform on-policy learning while maintaining the belief model. 

We test the two extensions on Atari 2600 games. The setup of the environments and the hyper-parameters remain the same with the previous experiments.

The experimental results in Table 2 show that the model-based version is far from good. Only in one game can it outperforms the baseline. It shows that by using pure model-based PPO, the approximation errors introduced by fitting the model can substantially affect the performance.

However, POME with non-decaying coefficient turns out to be not only good but even better than POME with decaying coefficient. It outperforms the original PPO in 33 games out of 49. This result indicates that even though adding the exploration value would increase the bias when estimating the advantage functions, it is empirically not harmful to the policy optimization algorithms in most of the environments. 

\begin{table}[!t]
\small
\caption{Comparison among PPO, POME with constant exploration coefficient, and the model-based extension of PPO. }
\begin{center}
\begin{tabular}{l|r|r|r}
{\sc Games} & {\sc PPO} & {\sc PPO} & {\sc POME}\\
& & {model-based} & {non-decay}\\
\hline
Alien              &    {\bf1850.3} &         1386.4 &         1658.1 \\
Amidar             &          674.6 &           27.7 &     {\bf704.0} \\
Assault            &         4971.9 &          872.2 &    {\bf6211.5} \\
Asterix            &         4532.5 &         1606.2 &    {\bf7235.0} \\
Asteroids          &    {\bf2097.5} &         1456.8 &         1788.4 \\
Atlantis           & {2311815} & {\bf2864448} &      2030477 \\
BankHeist          &    {\bf1280.6} &          159.4 &         1245.8 \\
BattleZone         &   {\bf17366.7} &         2790.0 &        15313.3 \\
BeamRider          &         1590.0 &          448.9 &    {\bf1989.2} \\
Bowling            &           40.1 &           28.0 &      {\bf66.2} \\
Boxing             &      {\bf94.6} &           52.5 &           92.6 \\
Breakout           &          274.8 &           18.8 &     {\bf399.2} \\
Centipede          &    {\bf4386.4} &         3343.6 &         2684.7 \\
ChopperCommand     &         3516.3 &         1603.5 &    {\bf3886.3} \\
CrazyClimber       &       110202.0 &         3640.0 &  {\bf112166.3} \\
DemonAttack        &        11378.4 &          169.4 &   {\bf18877.9} \\
DoubleDunk         &          -14.9 &          -17.9 &      {\bf-8.8} \\
Enduro             &          758.3 &           92.2 &     {\bf862.8} \\
FishingDerby       &           17.8 &          -62.7 &      {\bf19.3} \\
Freeway            &           32.5 &            4.0 &      {\bf33.0} \\
Frostbite          &     {\bf314.2} &          265.4 &          275.1 \\
Gopher             &         2932.9 &          102.4 &    {\bf5050.4} \\
Gravitar           &          737.2 &          150.0 &     {\bf773.8} \\
IceHockey          &      {\bf-4.2} &           -4.5 &           -4.5 \\
Jamesbond          &          560.7 &           37.2 &     {\bf871.8} \\
Kangaroo           &    {\bf9928.7} &         1422.0 &         2237.0 \\
Krull              &         7942.3 &         6091.0 &    {\bf8795.4} \\
KungFuMaster       &        23310.3 &        12255.5 &   {\bf27667.0} \\
MontezumaRevenge   &      {\bf42.0} &            0.0 &            0.0 \\
MsPacman           &         2096.5 &         2059.7 &    {\bf2101.0} \\
NameThisGame       &    {\bf6254.9} &         4485.1 &         5462.9 \\
Pitfall            &          -32.9 &         -591.5 &      {\bf-4.4} \\
Pong               &           20.7 &           16.1 &      {\bf20.8} \\
PrivateEye         &           69.5 &           39.3 &      {\bf98.9} \\
Qbert              &        14293.3 &         3308.2 &   {\bf14373.7} \\
Riverraid          &    {\bf8393.6} &         3646.9 &         8358.3 \\
RoadRunner         &        25076.0 &         4534.5 &   {\bf41740.7} \\
Robotank           &            5.5 &            2.6 &      {\bf13.3} \\
Seaquest           &         1204.5 &          592.4 &    {\bf1805.7} \\
SpaceInvaders      &          942.5 &          355.4 &     {\bf971.6} \\
StarGunner         &        32689.0 &         1497.5 &   {\bf33836.0} \\
Tennis             &          -14.8 &          -22.6 &     {\bf-12.6} \\
TimePilot          &    {\bf4232.0} &         3150.0 &         3970.7 \\
Tutankham          &     {\bf254.4} &           99.0 &          159.6 \\
UpNDown            &        95445.0 &         1589.4 &  {\bf203080.4} \\
Venture            &            0.0 &            0.0 &      {\bf55.0} \\
VideoPinball       &   {\bf37389.0} &        19720.4 &        31462.0 \\
WizardOfWor        &         4185.3 &         1551.0 &    {\bf6050.7} \\
Zaxxon             &         5008.7 &            1.5 &    {\bf6088.7} \\
\end{tabular}
\end{center}
\end{table}

\section{Conclusion and Discussion}
Due to the challenge of the trade-off between exploration and exploitation in environments with continuous state space, in this paper, we propose a novel policy-based algorithm named POME, which uses the discrepancy between both target values of model-free and model-based methods to build a relative measure of the exploration value. POME uses several practical techniques to enable the exploration while being stable, i.e., the clipped and centralized exploration value. In the actual algorithm, POME builds on the model-free PPO algorithm and adds the exploration bonus to the estimation of the advantage function. Experiments show that POME outperforms the original PPO in 33 Atari games out of 49.

There is yet a limitation that, if the reward signal is extremely sparse, the error of the target values would be close to $0$. So POME would have little improvement in such situations. For the future work, it would be interesting to further extend the insight to address the exploration problem in the environments with sparse reward signals, for example, by incorporating with some reward-independent curiosity-based exploration methods. 

\small\selectfont\bibliography{pome-bib}
\bibliographystyle{aaai}
\end{document}